\documentclass[11pt]{article}

\usepackage[final]{acl}

\usepackage{times}
\usepackage{latexsym}

\usepackage[T1]{fontenc}

\usepackage[utf8]{inputenc}

\usepackage{microtype}

\usepackage{inconsolata}

\usepackage{booktabs}

\usepackage{graphicx}
\graphicspath{{figures/}}

\usepackage{multirow}

\usepackage{float}

\title{Linguistic Blind Spots in Clinical Decision Extraction}

\author{Mohamed Elgaar \and Hadi Amiri \\
  University of Massachusetts Lowell\\
  \texttt{\{melgaar,hadi\}@cs.uml.edu}}

\begin{document}
\maketitle
\begin{abstract}
Extracting medical decisions from clinical notes is a key step for clinical decision support and patient-facing care summaries. 
We study how the linguistic characteristics of clinical decisions vary across decision categories and whether these differences explain extraction failures. Using MedDec discharge summaries annotated with decision categories from the Decision Identification and Classification Taxonomy for Use in Medicine (DICTUM), we compute seven linguistic indices for each decision span and analyze span-level extraction recall of a standard transformer model. We find clear category-specific signatures: 
drug-related and problem-defining decisions are entity-dense and telegraphic, whereas advice and precaution decisions contain more narrative, with higher stopword and pronoun proportions and more frequent hedging and negation cues. On the validation split, exact-match recall is 48\%, with large gaps across linguistic strata: recall drops from 58\% to 24\% from the lowest to highest stopword-proportion bins, and spans containing hedging or negation cues are less likely to be recovered. Under a relaxed overlap-based match criterion, recall increases to 71\%, indicating that many errors are span boundary disagreements rather than complete misses. Overall, narrative-style spans--common in advice and precaution decisions--are a consistent blind spot under exact matching, suggesting that downstream systems should incorporate boundary-tolerant evaluation and extraction strategies for clinical decisions.
\end{abstract}

\section{Introduction}

Medical decisions documented in clinical notes directly shape patient care and outcomes. Automatically extracting and classifying these decisions enables clinical decision support and patient-facing summaries~\citep{elgaar-etal-2024-meddec,elgaar-etal-2025-meddecxtract}, and can support efforts to identify and mitigate health disparities~\citep{Amiri2024.07.11.24310289}. However, decision spans differ systematically in linguistic form across decision types, which can create predictable failure modes for extraction models.

Clinical documentation is produced under time pressure, leading to a mix of shorthand fragments, list-like structures, and domain-specific terminology~\citep{sager-etal-1994-nlp}.  This leads to a strong stylistic heterogeneity across decision categories. Drug-related decisions, for example, tend to be entity-dense with medication names and dosages, while advice and precaution decisions are more narrative and instruction-like, often expressed with conditionals, hedging, and negation. This heterogeneity motivates two research questions: (\textbf{RQ1}) how do linguistic characteristics differ across medical decision categories, and (\textbf{RQ2}) are these characteristics associated with extraction errors?

We study these questions using the MedDec dataset~\citep{elgaar-etal-2024-meddec}, which contains discharge summaries annotated with decision spans according to the Decision Identification and Classification Taxonomy for Use in Medicine (DICTUM)~\citep{ofstad-etal-2016-dictum}. For each annotated span, we compute seven linguistic indices capturing readability, lexical composition, and discourse markers. Our results indicate that decision categories show distinct linguistic profiles and that recall (of a baseline transformer model) degrades substantially for spans with higher proportions of stopwords, hedging cues, and negation cues.

The contributions of this work are threefold. First, we characterize linguistic variation across nine medical decision categories in discharge summaries. Second, we link linguistic features to extraction failures at span level, including a 34-point recall drop between the lowest and highest stopword strata on validation data. Third, we connect these error patterns to downstream use: decision categories that are especially important for patient understanding disproportionately show the linguistic features most associated with extraction failures.\looseness-1

\section{Related Work}

Medical decision extraction builds on foundational clinical NLP resources and the DICTUM taxonomy. \citet{ofstad-etal-2016-dictum} introduced DICTUM to categorize clinical decisions into ten types (e.g., drug-related, therapeutic procedure, advice and precaution, treatment goals). MedDec adapts this taxonomy for computational extraction from discharge summaries~\citep{elgaar-etal-2024-meddec}; following the evaluation setup used throughout, our analyses restrict attention to nine categories, excluding legal/insurance due to limited annotation availability. 
\citet{Amiri2024.07.11.24310289} analyzed these decision spans and showed that the frequency of documented decisions differs significantly by patients’ language proficiency (and, to a lesser extent, sex) across decision types, indicating documentation disparities that can propagate into downstream clinical NLP models.\looseness-1

A parallel line of work studies the linguistic complexity of clinical text and argues for domain-specific measures beyond generic readability formulas. \citet{zheng-yu-2017-readability} showed that standard readability metrics correlate poorly with perceived difficulty in electronic health records, motivating domain-specific complexity modeling. \citet{crossley-etal-2020-eclippse} developed NLP-based readability predictors for physician messages, and \citet{jiang-xu-2024-medreadme} provided fine-grained medical sentence readability annotations, finding that jargon span density better matches human judgments.

Beyond difficulty estimation, prior work has highlighted that specific linguistic features can shape downstream behavior and that feature choice matters. \citet{brown-etal-2021-challenges} discussed challenges in measuring health literacy from patient-provider messages, emphasizing careful selection and interpretation of linguistic indices. Work on uncertainty phenomena in biomedical text indicates that negation, hedging, and related markers can affect extraction outcomes~\citep{nickels-etal-2024-uncertainty}. However, limited work has systematically quantified---within medical decision extraction---how linguistic properties of the target spans relate to extraction performance at the span level, which this paper addresses.\looseness-1

\begin{figure}[t]
\centering
\includegraphics[width=\columnwidth]{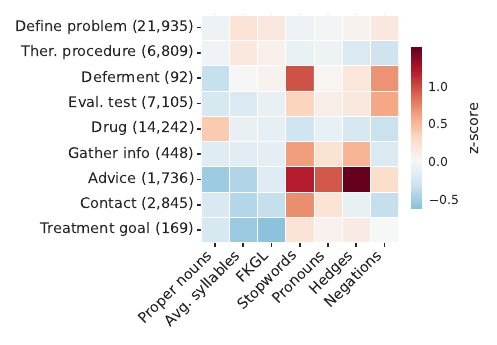}
\vspace{-25pt}
\caption{Z-scored mean linguistic indices by decision category. For each index, values are z-scored across all plotted spans as \(z=(x-\mu)/\sigma\) before computing category means. Red indicates above-average values; blue indicates below-average. Advice and precaution decisions show elevated hedging, negation, and pronoun use.\looseness-1}
\label{fig:heatmap}
\end{figure}

\section{Methods}

\subsection{MedDec Dataset}

MedDec~\citep{elgaar-etal-2024-meddec} contains 451 discharge summaries from MIMIC-III~\citep{johnson-etal-2016-mimic} annotated with medical decision spans. Spans are labeled with several categories: defining problem, evaluating test result, drug-related, therapeutic procedure related, gathering information, advice and precaution, contact-related, treatment goal, and deferment. The data contains 350/53/48 discharge summaries with 42,223/6,851/6,307 spans in train/validation/test splits. Corpus analyses (RQ1) use all spans across the corpus, while extraction reliability analyses (RQ2) uses validation data to avoid post-hoc selection on test and prevent test-set leakage. 

\begin{figure*}[t]
\centering
\includegraphics[width=0.95\textwidth]{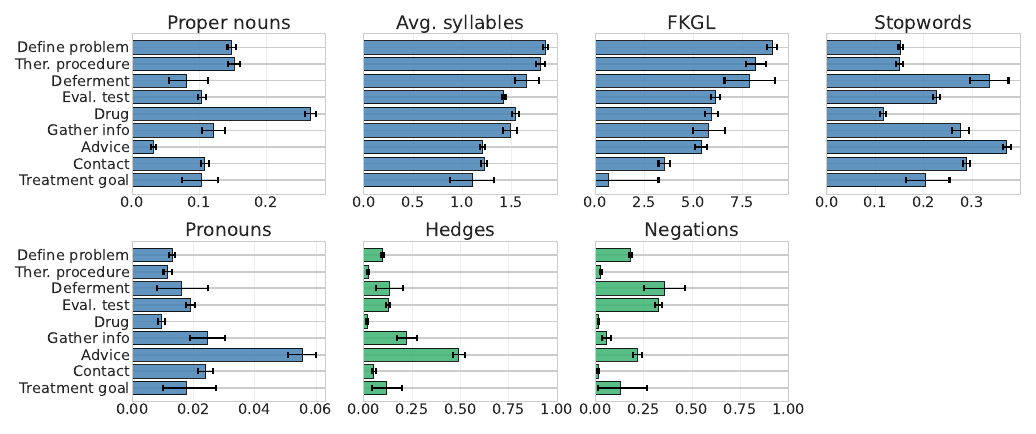}
\caption{Linguistic index distributions by decision category. Error bars indicate 95\% cluster-bootstrap confidence intervals. Categories are ordered by mean FKGL. Binary indices (hedge/negation presence) report the proportion of spans containing the marker.}
\label{fig:smallmultiples}
\end{figure*}

\subsection{Linguistic Indices}

Seven linguistic indices are computed for each gold span, extracted using the LFTK library~\citep{lee-lee-2023-lftk} and lexicon-based augmentation. The indices are grounded in prior research on clinical text complexity. Flesch-Kincaid Grade Level (FKGL)~\citep{kincaid-etal-1975-derivation} measures readability based on sentence and word length; prior work has examined its applicability to clinical documentation~\citep{zheng-yu-2017-readability}. Average syllables per word serves as a proxy for lexical complexity, reflecting the polysyllabic nature of medical terminology~\citep{sager-etal-1994-nlp}. The proportion of proper nouns captures entity density, as clinical text relies heavily on specific drug names, anatomical terms, and eponymous conditions. The proportion of stopwords indicates telegraphic versus narrative style; the exact stopword list used in this work is provided in Appendix~\ref{sec:stopwords}. The proportion of pronouns reflects referential language use. Binary indices include the presence of hedge and negation cues, computed using explicit cue lists (Appendix~\ref{sec:lexicons}).

\subsection{Extraction Model and Evaluation}

A RoBERTa-based~\citep{liu-etal-2019-roberta} span extraction model is trained on the MedDec training split following the standard configuration from \citet{elgaar-etal-2024-meddec}. We do not introduce additional model variants in this work; the goal is to analyze span-level reliability under a fixed, standard extractor. Accordingly, we keep the baseline configuration fixed and do not perform additional hyperparameter tuning in this study. For evaluation, each gold span is marked as matched (\texttt{is\_matched}$=1$) if the model produces a prediction with the correct category label and an exact match on normalized span text; otherwise \texttt{is\_matched}$=0$. As a sensitivity analysis, a relaxed match criterion marks a gold span as matched if there exists a predicted span of the same category with token-level overlap IoU \(\ge 0.5\) on token index ranges. Because partial overlaps can count as matches, improvements under IoU mainly indicate span-boundary disagreements rather than complete misses.

We report recall because the evaluation is gold-span--centric: each gold span contributes one \texttt{is\_matched} outcome, and averaging this indicator within a subset of gold spans yields \(TP/(TP+FN)\). Precision would require a complementary predicted-span--centric analysis to account for spurious predictions that do not match any gold span, which is outside the scope of this diagnostic reliability study.

To assess the relationship between linguistic indices and extraction performance, spans are stratified by index values. Continuous indices are divided into quantile bins (up to five bins); for indices with many tied values, quantile binning may yield fewer than five non-empty bins. Binary indices use their natural 0/1 values. Within each stratum, mean recall (proportion of matched spans) is computed with 95\% CI using cluster bootstrap resampling by document to account for within-document dependence. Appendix~\ref{sec:eval_details} reports bin ranges and sample sizes.

\begin{figure*}[t]
\centering
\includegraphics[width=0.85\textwidth]{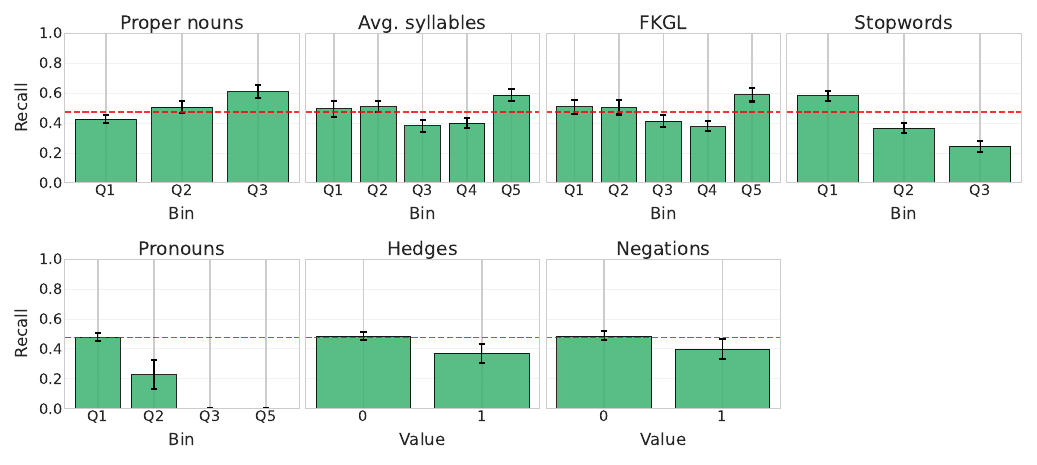}
\vspace{-10pt}
\caption{Span-level extraction recall by linguistic index bins (exact match). Dashed line indicates overall recall (48\%). Error bars show 95\% cluster-bootstrap confidence intervals. Higher stopword proportions are strongly associated with lower recall.}
\label{fig:recall}
\end{figure*}

\section{Results}

\noindent We answer RQ1 by characterizing how decision categories differ in linguistic indices (Figure~\ref{fig:heatmap}--\ref{fig:smallmultiples}). We answer RQ2 by measuring how these indices relate to span-level extraction recall on the validation split (Figure~\ref{fig:recall}--\ref{fig:recall_relaxed}, Table~\ref{tab:category_control}).

\subsection{Linguistic Variation Across Categories}

(RQ1) Figure~\ref{fig:heatmap} presents z-scored means of linguistic indices across decision categories. The analysis reveals distinct linguistic signatures. Drug-related decisions (n=14,242) exhibit high proper noun density, reflecting medication names, and low stopword proportions consistent with telegraphic documentation. Defining problem decisions (n=21,935) show similar patterns with elevated syllable counts indicating complex medical terminology.

In contrast, advice and precaution decisions (n=1,736) display markedly different characteristics: elevated stopwords, pronouns, hedges, and negations. This pattern reflects the narrative, conditional nature of patient instructions (e.g., ``If you experience chest pain, you should call your doctor''). Contact-related and treatment goal decisions similarly show higher stopword proportions and lower entity density, consistent with their communicative rather than diagnostic function.

Figure~\ref{fig:smallmultiples} presents the absolute distributions of each linguistic index across categories. Drug-related decisions show the highest proper noun density, while advice and precaution decisions show the lowest. Hedges are rare in drug-related decisions and common in advice and precaution decisions. These patterns suggest that decision categories occupy different regions of the linguistic feature space.

\begin{table}[t]
\centering
\small
\setlength{\tabcolsep}{4pt}
\begin{tabular}{lrr}
\toprule
Category & $n$ (val) & Recall (95\% CI) \\
\midrule
Defining problem & 2720 & 0.560 [0.533, 0.593] \\
Drug-related & 1954 & 0.545 [0.486, 0.597] \\
Advice and precaution & 205 & 0.478 [0.369, 0.578] \\
Therapeutic procedure related & 816 & 0.333 [0.284, 0.379] \\
Evaluating test result & 780 & 0.276 [0.225, 0.332] \\
Contact-related & 284 & 0.275 [0.226, 0.328] \\
Treatment goal & 46 & 0.130 [0.031, 0.474] \\
Gathering information & 41 & 0.000 [0.000, 0.000] \\
Deferment & 5 & 0.000 [0.000, 0.000] \\
\bottomrule
\end{tabular}
\caption{Category-stratified extraction recall on the validation split (exact match). Confidence intervals are cluster-bootstrapped by document.}
\label{tab:category_recall}
\end{table}

\subsection{Extraction Recall by Linguistic Features}

(RQ2) Figure~\ref{fig:recall} presents validation-set span-level recall stratified by linguistic index bins. Overall model recall is 48\%. Recall also varies substantially by decision category (Table~\ref{tab:category_recall}), with higher recall for entity-dense categories (e.g., defining problem, drug-related) and lower recall for several communicative or lower-frequency categories.

\begin{table}[t]
\centering
\small
\setlength{\tabcolsep}{4pt}
\begin{tabular}{lrr}
\toprule
Index (z-scored) & Coef. & 95\% CI \\
\midrule
Proportion of proper nouns & 0.051 & [-0.039, 0.144] \\
Average syllables & 0.060 & [-0.248, 0.375] \\
FKGL (Readability) & 0.017 & [-0.282, 0.296] \\
Proportion of stopwords & -0.556 & [-0.656, -0.471] \\
Proportion of pronouns & -0.084 & [-0.158, -0.013] \\
Hedge presence & -0.039 & [-0.103, 0.020] \\
Negation presence & 0.058 & [-0.026, 0.152] \\
\bottomrule
\end{tabular}
\caption{Category-controlled logistic regression of \texttt{is\_matched} on linguistic indices with category fixed effects. Coefficients are log-odds; confidence intervals are cluster-bootstrapped by document.}
\label{tab:category_control}
\end{table}

\begin{figure*}[t]
\centering
\includegraphics[width=0.8\textwidth]{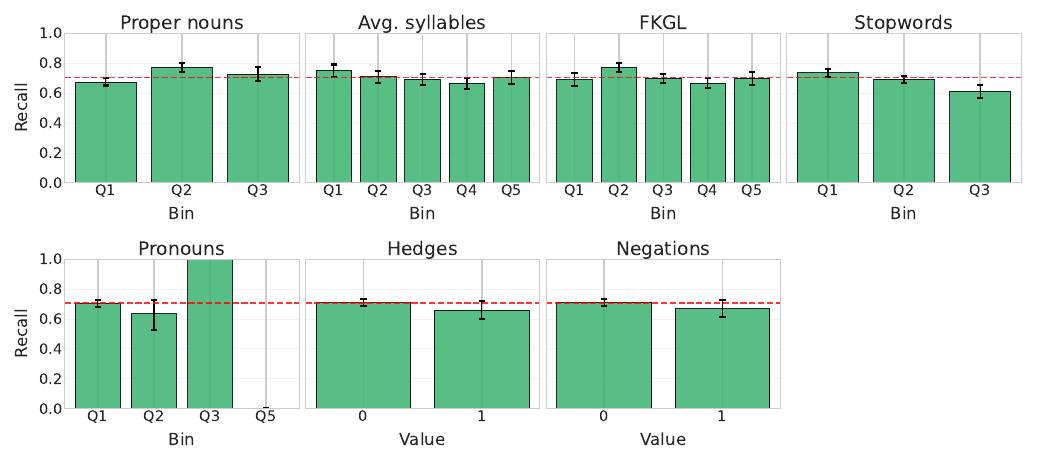}
\vspace{-10pt}
\caption{Span-level extraction recall by linguistic index bins under a relaxed overlap-based match criterion (IoU \(\ge 0.5\) within category). Dashed line indicates overall recall (71\%).}
\label{fig:recall_relaxed}
\end{figure*}

The strongest feature effect appears for stopwords: spans in the lowest stopword bin (telegraphic text) achieve 58\% recall, while those in the highest bin achieve 24\% recall. This 34 percentage point gap indicates that narrative-style spans pose substantial extraction challenges.

Proper noun density shows the opposite pattern: spans with the highest proper noun proportions achieve 61\% recall compared to 43\% for the lowest bin. This suggests that entity-dense spans (e.g., with specific drug names or anatomical terms) are associated with higher recall, potentially because they provide stronger lexical anchors.

Binary discourse markers are also associated with extraction difficulty. Spans containing hedging cues show 37\% recall compared to 48\% for spans without hedges. Similarly, spans with negation markers achieve 40\% recall versus 49\% for those without. These differences, while smaller in magnitude than the stopword effect, indicate that uncertainty and polarity markers correlate with extraction failures.

Controlling for category via logistic regression corroborates the stopword effect. Stopword proportion remains strongly negatively associated with matching probability (\(\beta=-0.56\), 95\% CI \([-0.66,-0.47]\)), while hedge and negation are not robust after controlling for category (Table~\ref{tab:category_control}).

Under the relaxed match criterion (IoU \(\ge 0.5\) within category), overall recall increases to 71\% and the qualitative feature gradients persist but shrink, e.g., the stopword gap decreases from 34 points (exact match) to 12 points (Figure~\ref{fig:recall_relaxed}).\looseness-1

\subsection{Implications}

Taken together, the corpus analysis (RQ1) and reliability results (RQ2) suggest a mismatch between instruction-like, patient-directed spans and what the model extracts reliably under exact match. Advice and precaution decisions are often patient-facing and more narrative in style, with higher stopword and pronoun proportions and more frequent uncertainty markers. Because stopword proportion is negatively associated with exact-match recall even after controlling for category, the baseline extractor is less reliable on instruction-like spans; deployments could add additional verification for advice and precaution or other narrative-style outputs.\looseness-1

\section{Discussion}

Span-level recall decreases as decision text becomes more narrative (e.g., higher stopword proportion), which the baseline extractor handles less reliably than telegraphic spans. The telegraphic, entity-dense style characteristic of drug-related and diagnostic decisions aligns with patterns that the extractor handles well. However, the narrative style of patient-directed advice, which often employs conditional language, hedging, and personal pronouns, is more difficult under exact match.

One plausible explanation is distributional: the model sees many entity-heavy spans and fewer advice-style constructions during training. Hedging and negation co-occur with advice-style spans, but their associations attenuate after controlling for category, suggesting these cues largely act through category/style composition.

Appendix~\ref{sec:demographics} reports demographic differences in the linguistic indices used here. Decision spans for non-English-speaking patients show lower lexical complexity compared to English-speaking patients, while documentation for male patients shows higher entity density than for female patients. This motivates evaluating extraction performance by demographic strata in future work.

For deployment, reliability signals could incorporate simple style indicators such as stopword proportion. For clinical decision support, confidence calibration can condition on category and span style (e.g., stopword-heavy vs.\ entity-dense spans). For patient-facing applications, advice and precaution extractions warrant targeted improvements or routine human review.

\section{Conclusion}

This study shows that the linguistic form of medical decisions is strongly associated with extraction model performance. Decision categories have distinct linguistic profiles, and exact-match recall drops by 34 points between the lowest and highest stopword strata on validation data. Narrative-style spans--especially in advice and precaution--perform worse under exact match; feature-stratified evaluation can help target improvements for these spans.

\section*{Limitations}

We evaluate one span-extraction architecture on discharge summaries from a single dataset; the observed associations between linguistic indices and extraction reliability may differ for other model families, training protocols, or clinical note types. Second, the evaluation is centered on recall over gold spans. While this design directly supports diagnosing which gold spans are missed, it does not quantify precision or the extent of spurious predictions, which would require an additional predicted-span--centric analysis. Third, the linguistic indices are surface-level operationalizations of complexity (e.g., cue-list matches for hedging and negation) and do not model scope or syntax; richer linguistic representations may yield more nuanced explanations of failure modes.

\bibliography{anthology_part1,anthology_part2,custom}

\clearpage
\appendix

\begin{figure*}[t]
\centering
\includegraphics[width=0.90\textwidth]{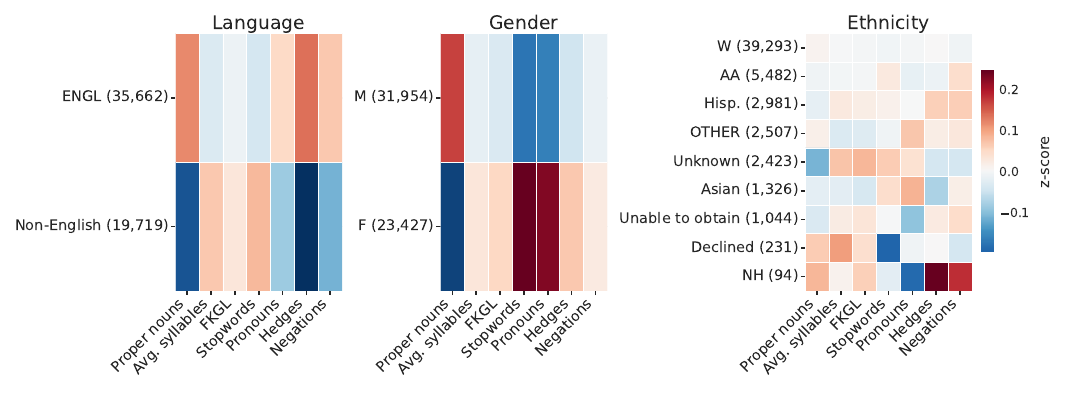}
\caption{Z-scored linguistic indices by patient demographics. Documentation for non-English-speaking and female patients shows lower entity density and more narrative style.}
\label{fig:demographics}
\end{figure*}

\section{Demographic Variation in Documentation Style}
\label{sec:demographics}

For the demographic stratification discussed in the main text, we provide summary statistics for each linguistic index. We report z-scored means per linguistic index within each demographic group, which puts indices on a common scale.

Figure~\ref{fig:demographics} presents z-scored linguistic indices stratified by patient demographics. Documentation for non-English-speaking patients (n=19,719) shows systematically lower lexical complexity---fewer proper nouns, fewer syllables per word, and lower FKGL---compared to documentation for English-speaking patients (n=35,662). This pattern may reflect linguistic accommodation by clinicians or differences in documentation detail.

Gender differences are also apparent. Documentation for male patients (n=31,954) shows higher proper noun density and syllable counts with fewer pronouns, suggesting a more formal, entity-dense style. Documentation for female patients (n=23,427) exhibits the opposite pattern with more pronouns and slightly more hedging, indicating a more narrative style. Major ethnic groups show relatively weak deviations, though small-sample groups exhibit extreme patterns likely attributable to sample size rather than systematic differences.

These demographic variations intersect with the extraction reliability findings: if documentation for certain patient groups exhibits more narrative-style language, extraction models may show differential performance across populations.

\begin{table}[htbp]
\centering
\small
\setlength{\tabcolsep}{4pt}
\begin{tabular}{llrr}
\toprule
Index & Bin & Range & $n$ (val) \\
\midrule
\multirow{3}{*}{Proper nouns} & 0 & [0.000, 0.130] & 4112 \\
 & 1 & [0.132, 0.333] & 1540 \\
 & 2 & [0.341, 1.000] & 1199 \\
\midrule
\multirow{5}{*}{Avg. syllables} & 0 & [0.071, 0.909] & 1375 \\
 & 1 & [0.913, 1.250] & 1481 \\
 & 2 & [1.257, 1.600] & 1265 \\
 & 3 & [1.605, 2.167] & 1360 \\
 & 4 & [2.176, 11.000] & 1370 \\
\midrule
\multirow{5}{*}{FKGL} & 0 & [-12.524, -1.450] & 1404 \\
 & 1 & [-1.443, 2.890] & 1344 \\
 & 2 & [2.898, 7.567] & 1364 \\
 & 3 & [7.570, 13.113] & 1453 \\
 & 4 & [13.158, 114.600] & 1286 \\
\midrule
\multirow{3}{*}{Stopwords} & 0 & [0.000, 0.167] & 4114 \\
 & 1 & [0.170, 0.333] & 1624 \\
 & 2 & [0.341, 1.000] & 1113 \\
\midrule
\multirow{4}{*}{Pronouns} & 0 & [0.000, 0.200] & 6777 \\
 & 1 & [0.211, 0.333] & 71 \\
 & 2 & [0.500, 0.500] & 2 \\
 & 3 & [1.000, 1.000] & 1 \\
\midrule
Hedges & 0/1 & \{0,1\} & 6314 / 537 \\
Negations & 0/1 & \{0,1\} & 6013 / 838 \\
\bottomrule
\end{tabular}
\caption{Quantile-bin ranges and sample sizes used for stratified recall estimates on the validation split. Some indices yield fewer than five bins due to tied values in quantile binning.}
\label{tab:bin_edges}
\end{table}

\section{Additional Evaluation Details}

This appendix documents the evaluation protocol and the stratified analyses referenced in the main text. It specifies (i) the decision categories included, (ii) how spans are counted within strata, and (iii) how uncertainty is estimated, so readers can reproduce the reported trends. Tables~\ref{tab:bin_edges}--\ref{tab:category_control} cover the binning/control details; Figure~\ref{fig:recall_relaxed} repeats the stratified analysis under a relaxed match rule.

We use the same span-level definition as in the main analysis. Each gold span receives a binary \texttt{is\_matched} indicator under the stated match criterion, and recall is the mean of this indicator within the relevant subset of spans.

We compute confidence intervals with a document-level cluster bootstrap. This prevents multiple spans from the same note from making uncertainty look too small.

\subsection{Category-stratified recall and binning details}
\label{sec:eval_details}
Table~\ref{tab:category_recall} (main text) reports category-stratified exact-match recall on the validation split. Unless noted otherwise, figures and tables use nine DICTUM-aligned decision categories; we exclude the legal/insurance category because it is out of scope for the evaluation setup used here.

Table~\ref{tab:bin_edges} lists the bin ranges and sample sizes used for stratified recall estimates. These bins define the strata used in the stratified recall plots. Quantile binning is performed per index on the validation split; for indices with many tied values, quantile binning can collapse to fewer than five bins. Accordingly, some indices yield three (or fewer) non-empty bins even when the target is five quantiles.

The tables above summarize stratified recall directly; the regression in the main text (Table~\ref{tab:category_control}) tests whether the same patterns hold after adjusting for differences in category composition. Coefficients are reported in log-odds. Since indices are z-scored before fitting, coefficients can be compared in magnitude across indices.

\subsection{Relaxed match sensitivity analysis}
To separate full misses from span-boundary disagreements, Figure~\ref{fig:recall_relaxed} repeats the stratified analysis under a relaxed overlap-based match criterion (IoU \(\ge 0.5\) within category). Recall increases under IoU \(\ge 0.5\), but the direction of the stratified trends is similar to the exact-match results. This suggests boundary disagreements explain some of the errors, while full misses remain.

The relaxed-match analysis helps distinguish segmentation errors from category confusion. When recall improves under IoU-based matching, the model often finds the correct decision content but disagrees on exact boundaries.

\section{Lexical Resources for Indices}
\label{sec:lexical_resources}

\subsection{Cue lists for hedges and negation}
\label{sec:lexicons}

\begin{table}[H]
\centering
\small
\begin{tabular}{llll}
\toprule
\multicolumn{4}{c}{Negation cues} \\
\midrule
absent & cannot & denied & denies \\
deny & free of & lack of & lacking \\
lacks & neg & negative & neither \\
never & no & no evidence & no signs \\
no symptoms & nobody & non & none \\
nor & not & nothing & nowhere \\
rule out & ruled out & rules out & unable \\
unlikely & unremarkable & without & \\
\bottomrule
\end{tabular}
\caption{Lexicon of negation cues used for index computation.}
\label{tab:negation_cues}
\end{table}

Hedge and negation indicators are computed by matching span text against the cue sets in Tables~\ref{tab:negation_cues} and~\ref{tab:hedge_cues} (case-insensitive string match; multiword cues are matched as phrases). We use string-matched cue lists so the indices map directly to explicit cue strings in the span text.

These indicators are coarse: they do not model scope or syntactic attachment. We use them as surface markers of uncertainty and polarity inside extracted decision spans.

\begin{table}[t]
\centering
\small
\begin{tabular}{llll}
\toprule
\multicolumn{4}{c}{Hedge cues} \\
\midrule
about & appeared & appears & approximate \\
approximately & around & await & awaiting \\
concern & concerned & consider & considered \\
could & differential & equivocal & estimated \\
if & likely & may & might \\
or & pending & perhaps & possible \\
possibly & potential & potentially & presumed \\
presumptive & probable & probably & questionable \\
seems & seemed & should & suggested \\
suggests & suspect & suspected & suspicion \\
tentative & uncertain & unclear & unlikely \\
versus & vs & whether & \\
\bottomrule
\end{tabular}
\caption{Lexicon of hedge cues used for index computation.}
\label{tab:hedge_cues}
\end{table}

\subsection{Stopword list}
\label{sec:stopwords}

This appendix lists the exact stopword set used to compute Proportion of stopwords. We include the stopword list to make the stopword-based stratification reproducible and to specify what this study treats as narrative ``glue'' text. The set was generated by a short Python script and is based on spaCy's English stopwords.\footnote{\url{https://github.com/explosion/spaCy/blob/master/spacy/lang/en/stop_words.py}.}

We print the stopwords as a readable list; any duplicates in the printed table do not matter because the computation uses set membership.
\begin{table*}[t]
\centering
\footnotesize
\setlength{\tabcolsep}{2pt}
\renewcommand{\arraystretch}{0.9}
\begin{tabular}{llllllllll}
\toprule
\multicolumn{10}{c}{Stopwords} \\
\midrule
  a & about & above & across & after & afterwards & again & against & all & almost \\
  alone & along & already & also & although & always & am & among & amongst & amount \\
  an & and & another & any & anyhow & anyone & anything & anyway & anywhere & are \\
  around & as & at & back & be & became & because & become & becomes & becoming \\
  been & before & beforehand & behind & being & below & beside & besides & between & beyond \\
  both & bottom & but & by & ca & call & can & cannot & could & did \\
  do & does & doing & done & down & due & during & each & eight & either \\
  eleven & else & elsewhere & empty & enough & even & ever & every & everyone & everything \\
  everywhere & except & few & fifteen & fifty & first & five & for & former & formerly \\
  forty & four & from & front & full & further & get & give & go & had \\
  has & have & he & hence & her & here & hereafter & hereby & herein & hereupon \\
  hers & herself & him & himself & his & how & however & hundred & i & if \\
  in & indeed & into & is & it & its & itself & just & keep & last \\
  latter & latterly & least & less & made & make & many & may & me & meanwhile \\
  might & mine & more & moreover & most & mostly & move & much & must & my \\
  myself & n't & n't & n't & name & namely & neither & never & nevertheless & next \\
  nine & no & nobody & none & noone & nor & not & nothing & now & nowhere \\
  of & off & often & on & once & one & only & onto & or & other \\
  others & otherwise & our & ours & ourselves & out & over & own & part & per \\
  perhaps & please & put & quite & rather & re & really & regarding & s & same \\
  say & see & seem & seemed & seeming & seems & serious & several & she & should \\
  show & side & since & six & sixty & so & some & somehow & someone & something \\
  sometime & sometimes & somewhere & still & such & take & ten & than & that & the \\
  their & them & themselves & then & thence & there & thereafter & thereby & therefore & therein \\
  thereupon & these & they & third & this & those & though & three & through & throughout \\
  thru & thus & to & together & too & top & toward & towards & twelve & twenty \\
  two & under & until & up & unless & upon & us & used & using & various \\
  very & via & was & we & well & were & what & whatever & when & whence \\
  whenever & where & whereafter & whereas & whereby & wherein & whereupon & wherever & whether & which \\
  while & whither & who & whoever & whole & whom & whose & why & will & with \\
  within & without & would & yet & you & your & yours & yourself & yourselves & 'd \\
  'll & 'm & 're & 's & 've & & & & & \\
\bottomrule
\end{tabular}
\caption{Stopword list used to compute stopword proportion.}
\label{tab:stopwords}
\end{table*}

This list is intentionally broad: it includes function words, auxiliaries, and common discourse glue that increases span narrativity without necessarily adding domain content. Because the metric is computed as a proportion over tokens in the gold span, longer instructional spans with more connective language will naturally score higher than telegraphic problem and medication fragments.

\end{document}